# Recognition of Handwritten Digit using Convolutional Neural Network in Python with Tensorflow and Comparison of Performance for Various Hidden Layers

Fathma Siddique[1#], Shadman Sakib[2*], Md. Abu Bakr Siddique[2$]
[1]Department of CSE, International University of Business Agriculture and Technology, Dhaka 1230, Bangladesh
[2]Department of EEE, International University of Business Agriculture and Technology, Dhaka 1230, Bangladesh
siddiquefathma@gmail.com[#], sakibshadman15@gmail.com[*], absiddique@iubat.edu[$]

*Abstract*—In recent times, with the increase of Artificial Neural Network (ANN), deep learning has brought a dramatic twist in the field of machine learning by making it more artificially intelligent. Deep learning is remarkably used in vast ranges of fields because of its diverse range of applications such as surveillance, health, medicine, sports, robotics, drones, etc. In deep learning, Convolutional Neural Network (CNN) is at the center of spectacular advances that mixes Artificial Neural Network (ANN) and up to date deep learning strategies. It has been used broadly in pattern recognition, sentence classification, speech recognition, face recognition, text categorization, document analysis, scene, and handwritten digit recognition. The goal of this paper is to observe the variation of accuracies of CNN to classify handwritten digits using various numbers of hidden layers and epochs and to make the comparison between the accuracies. For this performance evaluation of CNN, we performed our experiment using Modified National Institute of Standards and Technology (MNIST) dataset. Further, the network is trained using stochastic gradient descent and the backpropagation algorithm.

## I. INTRODUCTION

With time the numbers of fields are increasing in which deep learning can be applied. In deep learning, Convolutional Neural Networking (CNN) [1, 2] is being used for visual imagery analyzing. Object detection, face recognition, robotics, video analysis, segmentation, pattern recognition, natural language processing, spam detection, topic categorization, regression analysis, speech recognition, image classification are some of the examples that can be done using Convolutional Neural Networking. The accuracies in these fields including handwritten digits recognition using Deep Convolutional Neural Networks (CNNs) have reached human level perfection. Mammalian visual systems' biological model is the one by which the architecture of the CNN is inspired. Cells in the cat's visual cortex are sensitized to a tiny area of the visual field identified which is recognized as the receptive field [3]. It was found by D. H. Hubel et al. in 1062. The neocognitron [4], the pattern recognition model inspired by the work of D. H. Hubel et al. [5, 6] was the first computer vision. It was introduced by Fukushima in 1980. In 1998, the framework of CNNs is designed by LeCun et al. [7] which had seven layers of convolutional neural networks. It was adept in handwritten digits classification direct from pixel values of images [8]. Gradient descent and back propagation algorithm [9] is used for training the model. In handwritten recognition digits, characters are given as input. The model can be recognized by the system. A simple artificial neural network (ANN) has an input layer, an output layer and some hidden layers between the input and output layer. CNN has a very similar architecture as ANN. There are several neurons in each layer in ANN. The weighted sum of all the neurons of a layer becomes the input of a neuron of the next layer adding a biased value. In CNN the layer has three dimensions. Here all the neurons are not fully connected. Instead, every neuron in the layer is connected to the local receptive field. A cost function generates in order to train the network. It compares the output of the network with the desired output. The signal propagates back to the system, again and again, to update the shared weights and biases in all the receptive fields to minimize the value of cost function which increases the network's performance [10-12]. The goal of this article is to observe the influence of hidden layers of a CNN for handwritten digits. We have applied a different type of Convolutional Neural Network algorithm on Modified National Institute of Standards and Technology (MNIST) dataset using Tensorflow, a Neural Network library written in python. The main purpose of this paper is to analyze the variation of outcome results for using a different combination of hidden layers of Convolutional Neural Network. Stochastic gradient and backpropagation algorithm are used for training the network and the forward algorithm is used for testing.

## II. LITERATURE REVIEW

CNN is playing an important role in many sectors like image processing. It has a powerful impact on many fields. Even, in nano-technologies like manufacturing semiconductors, CNN is used for fault detection and classification [13]. Handwritten digit recognition has become an issue of interest among researchers. There are a large number of papers and articles are being published these days about this topic. In research, it is shown that Deep Learning algorithm like multilayer CNN using Keras with Theano and Tensorflow gives the highest accuracy in comparison with the most widely used machine learning algorithms like SVM, KNN & RFC. Because of its highest accuracy, Convolutional Neural Network (CNN) is being used on a large scale in image classification, video analysis, etc. Many researchers are trying to make sentiment recognition in a sentence. CNN is being used in natural language processing and sentiment recognition by varying different parameters [14]. It is pretty challenging to get a good performance as more parameters are needed for the large-scale neural network. Many researchers are trying to increase the accuracy with less error in CNN. In another research, they have shown that deep nets perform better when they are trained by simple back-propagation. Their architecture results in the lowest error rate on MNIST compare to NORB and CIFAR10 [15]. Researchers are working on this issue to reduce the error rate as much as possible in handwriting recognition. In one research, an error rate of 1.19% is achieved using 3-NN trained and tested on MNIST [22]. Deep CNN can be adjustable with the input image noise [16]. Coherence recurrent convolutional network (CRCN) is a multimodal neural architecture [17]. It is being used in recovering sentences in an image. Some researchers are trying to come up with new techniques to avoid drawbacks of traditional convolutional layer's. Ncfm (No combination of feature maps) is a technique which can be applied for better performance using MNIST datasets [18]. Its accuracy is 99.81% and it can be applied for large-scale data. New applications of CNN are developing day by day with many kinds of research. Researchers are trying hard to minimize error rates. Using MNIST datasets and CIFAR, error rates are being observed [19]. To clean blur images CNN is being used. For this purpose, a new model was proposed using MNIST dataset. This approach reaches an accuracy of 98% and loss range 0.1% to 8.5% [20]. In Germany, a traffic sign recognition model of CNN is suggested. It proposed a faster performance with 99.65% accuracy [21]. Loss function was designed, which is applicable for light-weighted 1D and 2D CNN. In this case, the accuracies were 93% and 91% respectively [22, 23].

III. MODELING OF CONVOLUTIONAL NEURAL NETWORK TO CLASSIFY HANDWRITTEN DIGITS

To recognize the handwritten digits, a seven-layered convolutional neural network with one input layer followed by five hidden layers and one output layer is designed and illustrated in figure 1.

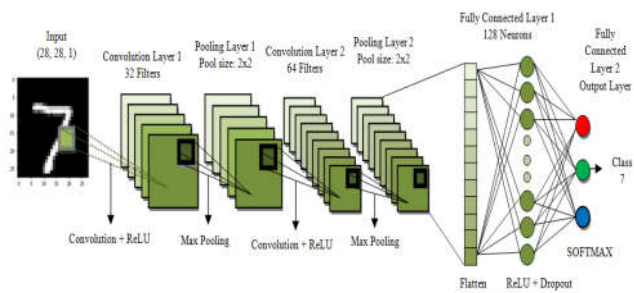

Fig. 1. A seven-layered convolutional neural network for digit recognition

The input layer consists of 28 by 28 pixel images which mean that the network contains 784 neurons as input data. The input pixels are grayscale with a value 0 for a white pixel and 1 for a black pixel. Here, this model of CNN has five hidden layers. The first hidden layer is the convolution layer 1 which is responsible for feature extraction from an input data. This layer performs convolution operation to small localized areas by convolving a filter with the previous layer. In addition, it consists of multiple feature maps with learnable kernels and rectified linear units (ReLU). The kernel size determines the locality of the filters. ReLU is used as an activation function at the end of each convolution layer as well as a fully connected layer to enhance the performance of the model. The next hidden layer is the pooling layer 1. It reduces the output information from the convolution layer and reduces the number of parameters and computational complexity of the model. The different types of pooling are max pooling, min pooling, average pooling, and L2 pooling. Here, max pooling is used to subsample the dimension of each feature map. Convolution layer 2 and pooling layer 2 which has the same function as convolution layer 1 and pooling layer 1 and operates in the same way except for their feature maps and kernel size varies. A Flatten layer is used after the pooling layer which converts the 2D featured map matrix to a 1D feature vector and allows the output to get handled by the fully connected layers. A fully connected layer is another hidden layer also known as the dense layer. It is similar to the hidden layer of Artificial Neural Networks (ANNs) but here it is fully connected and connects every neuron from the previous layer to the next layer. In order to reduce overfitting, dropout regularization method is used at fully connected layer 1. It randomly switches off some neurons during training to improve the performance of the network by making it more robust. This causes the network to become capable of better generalization and less compelling to overfit the training data. The output layer of the network consists of ten neurons and determines the digits numbered from 0 to 9. Since the output layer uses an activation function such as softmax, which is used to enhance the performance of the model, classifies the output digit from 0 through 9 which has the highest activation value.

The MNIST handwritten digits [24] database is used for the experiment. Out of 70,000 scanned images of handwritten digits from the MNIST database, 60,000 scanned images of digits are used for training the network and 10,000 scanned images of digits are used to test the network. The images that are used for training and testing the network all are the grayscale image with a size of 28×28 pixels. Character x is used to represent a training input where x is a 784-dimensional vector as the input of x is regarded as 28×28 pixels. The equivalent desired output is expressed by y(x), where y is a 10-dimensional vector. The network aims is to find the convenient weights and biases so that the output of the network approximates y(x) for all training inputs x as it completely depends on weight values and bias values. To compute the network performances, a cost function is defined, expressed by equation 1 [25].

$$C(w,b) = \frac{1}{2n} \sum_x \left[ y(x) - a^2 \right]^2 \qquad (1)$$

Where w is the cumulation of weights in the network, b is all the biases, n is the total number of training inputs and a is the actual output. The actual output a depends on x, w, and b. C(w,b) is non-negative as all the terms in the sum is non-negative. Moreover, C(w,b)=0, precisely when desired output y(x) is comparatively equal to the actual output, a, for all training inputs, n. To reduce the cost C(w,b) to a smaller degree as a function of weight and biases, the training algorithm has to find a set of weight and biases which cause the cost to become as small as possible. This is done using an algorithm known as gradient descent. In other words, gradient descent is an optimization algorithm that twists its parameters iteratively to minimize a cost function to its local minimum. The gradient descent algorithm deploys the following equations [25] to set the weight and biases.

$$w^{new} = w^{old} - \eta \frac{\partial C}{\partial w^{old}} \quad (2)$$

$$b^{new} = b^{old} - \eta \frac{\partial C}{\partial b^{old}} \quad (3)$$

And to attain the global minimum of the cost C(w,b) shown in figure 2.

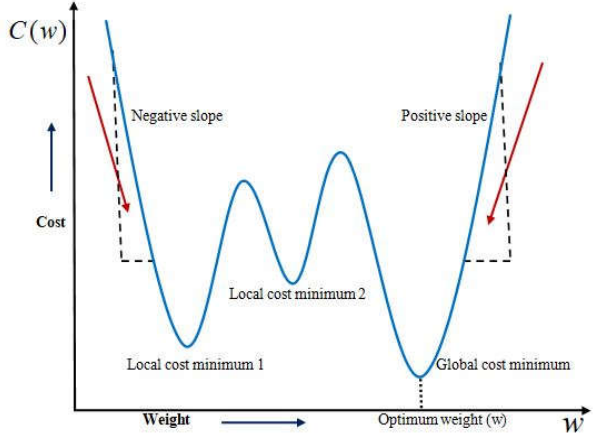

Fig. 2. Graphical Representation of Cost vs. Weight

However, the gradient descent algorithm may be unusable when the training data size is very large. Therefore, to enhance the performance of the network, a stochastic version of the algorithm is used. In Stochastic Gradient Descent (SDG) a small number of iteration will find effective solutions for the optimization problems. Moreover, in SDG, a small number of iteration will lead to a suitable solution. The Stochastic Gradient Descent algorithm utilizes the following equations [25]:

$$w^{new} = w^{old} - \frac{\eta}{m} \frac{\partial C_{xj}}{\partial w^{old}} \quad (4)$$

$$b^{new} = b^{old} - \frac{\eta}{m} \frac{\partial C_{xj}}{\partial w^{old}} \quad (5)$$

The output of the network can be expressed by:
$$a = f(z) = f(wa+b) \quad (6)$$

To find the amount of weight that contributes to the total error of the network Backpropagation method is used. The backpropagation of the network is illustrated by the following equations [25]:

$$\delta^L = \frac{\partial C}{\partial a^{(L)}} \frac{\partial a^{(L)}}{\partial z^{(L)}} = \frac{1}{n}(a^{(L)} - 1)f'(z^{(L)}) \quad (7)$$

$$\delta^l = \frac{\partial C}{\partial z^{(l)}} = \frac{\partial C}{\partial z^{(l)}} \frac{\partial z^{(l+1)}}{\partial z^{(l)}} = \frac{\partial z^{(l+1)}}{\partial z^{(l)}} \delta^{l+1} = w^{l+1}\delta^{l+1} f'(z^l) \quad (8)$$

$$\frac{\partial C}{\partial b^{(l)}} = \delta^l \quad (9)$$

$$\frac{\partial C}{\partial w^{(L)}} = a^{l-1}\delta^l \quad (10)$$

## IV. MNIST DATASET

Modified National Institute of Standards and Technology (MNIST) is a large set of computer vision dataset which is extensively used for training and testing different systems. It was created from the two special datasets of National Institute of Standards and Technology (NIST) which holds binary images of handwritten digits. The training set contains handwritten digits from 250 people, among them 50% training dataset was employees from the Census Bureau and the rest of it was from high school students [26]. However, it is often attributed as the first datasets among other datasets to prove the effectiveness of the neural networks.

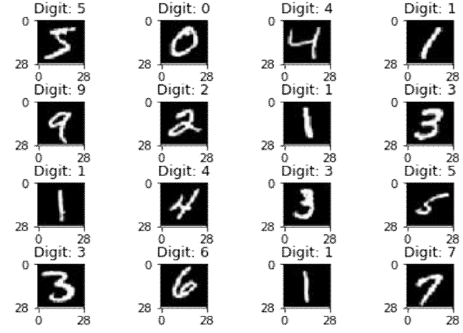

Fig. 3. Sample images of MNIST handwritten digit dataset

The database contains 60,000 images used for training as well as few of them can be used for cross-validation purposes and 10,000 images used for testing [27]. All the digits are grayscale and positioned in a fixed size where the intensity lies at the center of the image with 28×28 pixels. Since all the images are 28×28 pixels, it forms an array which can be flattened into 28*28=784 dimensional vector. Each component of the vector is a binary value which describes the intensity of the pixel.

## V. RESULTS AND DISCUSSION

### A. Discussion of the Obtained Simulated Results

In this section, CNN has been applied on the MNIST dataset in order to observe the variation of accuracies for handwritten digits. The accuracies are obtained using Tensorflow in python. Training and validation accuracy for 15 different epochs were observed exchanging the hidden layers for various combinations of convolution and hidden layers by taking the batch size 100 for all the cases. Figure 4, 5, 6, 7, 8, and 9 shows the performance of CNN for different combinations of convolution and hidden layers.

Table 1 shows the minimum and maximum training and validation accuracies of CNN found after the simulation for the six different cases by varying number of hidden layers for the recognition of handwritten digits.

TABLE I. PERFORMANCE OF CNN FOR THE SIX DIFFERENT CASES FOR VARIOUS HIDDEN LAYERS AND EPOCHS

| Case | Number of Hidden Layers | Batch Size | Minimum Training Accuracy | | Minimum Validation Accuracy | | Maximum Training Accuracy | | Maximum Validation Accuracy | | Overall Performance Validation Accuracy (%) |
|---|---|---|---|---|---|---|---|---|---|---|---|
| | | | Epoch | Accuracy (%) | Epoch | Accuracy (%) | Epoch | Accuracy (%) | Epoch | Accuracy (%) | |
| 1 | 3 | 100 | 1 | 91.94 | 1 | 97.73 | 13 | 98.99 | 14 | 99.16 | 99.11 |
| 2 | 4 | 100 | 1 | 90.11 | 1 | 97.74 | 14 | 98.94 | 14 | 99.24 | 99.21 |
| 3 | 3 | 100 | 1 | 94.35 | 3 | 98.33 | 15 | 100 | 15 | 99.06 | 99.06 |
| 4 | 4 | 100 | 1 | 92.94 | 1 | 97.79 | 15 | 99.92 | 13 | 99.92 | 99.20 |
| 5 | 3 | 100 | 1 | 91.80 | 1 | 98.16 | 13 | 99.09 | 12 | 99.12 | 99.09 |
| 6 | 4 | 100 | 1 | 90.50 | 1 | 97.13 | 15 | 99.24 | 13 | 99.26 | 99.07 |

In the first case shown in figure 4, the first hidden layer is the convolutional layer 1 which is used for the feature extraction. It consists of 32 filters with the kernel size of 3×3 pixels and the rectified linear units (ReLU) is used as an activation function to enhance the performance. The next hidden layer is the convolutional layer 2 consists of 64 filters with a kernel size of 3×3 pixels and ReLU. Next, a pooling layer 1 is defined where max pooling is used with a pool size of 2×2 pixels to minimize the spatial size of the output of a convolution layer. A regularization layer dropout is used next to the pooling layer 1 where it randomly eliminates 25% of the neurons in the layer to reduce overfitting. A flatten layer is used after the dropout which converts the 2D filter matrix into 1D feature vector before entering into the fully connected layers. The next hidden layer used after the flatten layer is the fully connected layer 1 consists of 128 neurons and ReLU. A dropout with a probability of 50% is used after the fully connected layer 1. Finally, the output layer which is used here as fully connected layer 2 contains 10 neurons for 10 classes and determines the digits numbered from 0 to 9.

A softmax activation function is incorporated with the output layer to output digit from 0 through 9. The CNN is fit over 15 epochs with a batch size of 100. The overall validation accuracy in the performance is found at 99.11%. At epoch 1 the minimum training accuracy of 91.94% is found and 97.73% of validation accuracy is found. At epoch 13, the maximum training accuracy is found 98.99% and at epoch 14, the maximum validation accuracy is found 99.16%. The total test loss for this case is found approximately 0.037045.

Figure 5 is defined for case 2, where convolution 1, pooling 1 and convolution 2, pooling 2 is used one after another. A dropout is used followed by the flatten layer and fully connected layer 1. Before the fully connected layer 2 another dropout is used. The dimensions and parameters used here and for the next cases are same which are used earlier for case 1. The overall validation accuracy in the performance is found 99.21%. At epoch 1 the minimum training and validation accuracy are found. The minimum training accuracy is 90.11% and the minimum validation accuracy is 97.74%. The maximum training and validation accuracy are found at epoch 14. The maximum training and validation accuracies are 98.94% and 99.24%. The total test loss is found approximately 0.026303.

For case 3, shown in figure 6, where two convolutions are taken one after another followed by a pooling layer. After the pooling layer, a flatten layer is used followed by the two fully connected layers without any dropout. The overall validation accuracy in the performance is found 99.06%. The minimum training accuracy is found 94.35% at epoch 1 and epoch 3, the minimum validation accuracy is found 98.33%. The maximum training and validation accuracies are 1% and 99.06% found at epoch 15. The total test loss is found approximately 0.049449.

Similarly, in case 4 shown in figure 7, convolution 1, pooling 1 and convolution 2, pooling 2 are used alternately followed by a flatten layer and two fully connected layers without any dropout. The overall validation accuracy in the performance is found 99.20%. At epoch 1 the minimum training and validation accuracy are found.

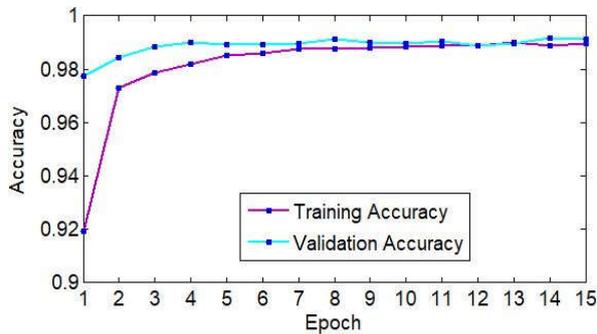

Fig. 4. Observed accuracy for case 1

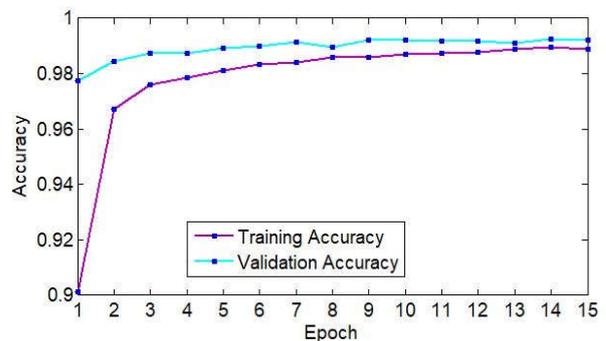

Fig. 5. Observed accuracy for case 2

The minimum training accuracy is 92.94% and the minimum validation accuracy is 97.79%. The maximum training accuracy is found 99.92% at epoch 15 and epoch 13, the maximum validation accuracy also found 99.92%. The total test loss is found approximately 0.032287.

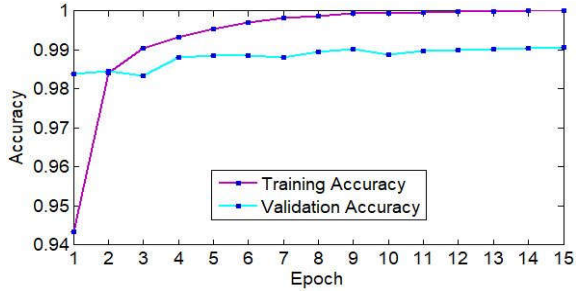
Fig. 6. Observed accuracy for case 3

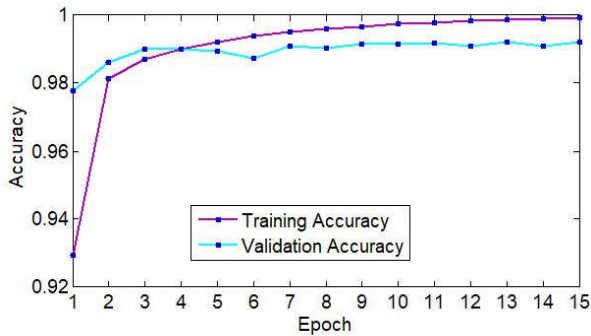
Fig. 7. Observed accuracy for case 4

Again, for case 5 shown in figure 8, two convolutions are used one after another followed by a pooling layer, flatten layer and fully connected layer 1. A dropout is used before the fully connected layer 2. The overall validation accuracy in the performance is found 99.09%. The minimum training and validation accuracy are found at epoch 1. The minimum training accuracy is 91.80% and the minimum validation accuracy is 98.16%. At epoch 13, the maximum training accuracy is found 99.09% and the maximum validation accuracy is found 99.12% at epoch 12. The total test loss is found approximately 0.034337.

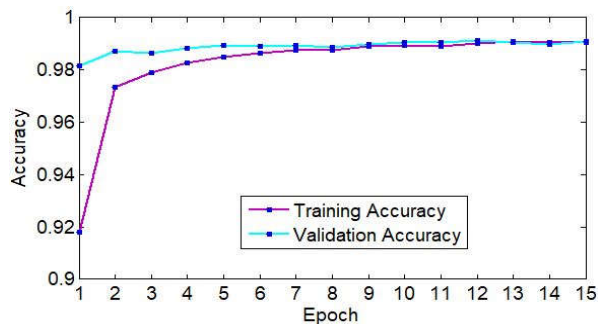
Fig. 8. Observed accuracy for case 5

Finally, for case 6 shown in figure 9, convolution 1, pooling 1 and convolution 2, pooling 2 are used alternately followed by a flatten layer and fully connected layer 1. A dropout is used before the fully connected layer 2. The overall validation accuracy in the performance is found 99.07%. At epoch 1 the minimum training and validation accuracy are found. The minimum training accuracy is 90.5% and the minimum validation accuracy is 97.13%. The maximum training accuracy is found 99.24% at epoch 15 and the maximum validation accuracy is found 99.26% at epoch 13. The total test loss is found approximately 0.028596.

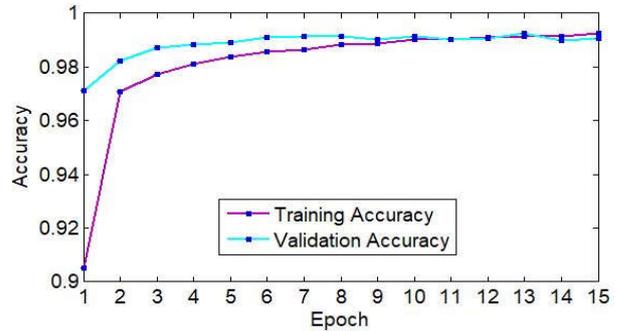
Fig. 9. Observed accuracy for case 6

*B. Comparison with Existing Research Work*

There are several methods of digit recognition. The handwritten digit recognition can be improved using some widely held methods of the neural network like Deep Neural Network (DNN), Deep Belief Network (DBF) and Convolutional Neural Network (CNN), etc.

Tavanaei et al. proposed multi-layered unsupervised learning in a spiking CNN model where they used MNIST dataset to clear the blur images and found the overall accuracy of 98% and the range of performance loss was 0.1% to 8.5% [20]. Rezoana et al. [28] proposed a seven-layered Convolutional Neural Network for the purpose of handwritten digit recognition where they used MNIST dataset to evaluate the impact of the pattern of the hidden layers of CNN over the performance of the overall network. They have plotted the loss curves against the number of epochs and found that the performance loss was below 0.1 for most of the cases and sometimes, in some cases, the loss was less than 0.05. In another paper, Siddique et al. [29] proposed an L-layered feed-forward neural network for the handwritten digit recognition where they have applied neural network with different layers on the MNIST dataset to observe the variation of accuracies of ANN for different combinations of hidden layers and epochs. Their maximum accuracy in the performance was found 97.32% for 4 hidden layers at 50 epochs.

Comparing with their above performances based on MNIST dataset for the purpose of digit recognition we have achieved better performance for the CNN. In our experiment, we have found the maximum training accuracy 100% and maximum validation accuracy 99.92% both at epoch 15. The overall performance of the network is found 99.21%. Moreover, the overall loss ranged from 0.026303 to 0.049449. Hence, this proposed method of CNN is more efficient than the other existing method for digit recognition.

VI. CONCLUSION

In this paper, the variations of accuracies for handwritten digit were observed for 15 epochs by varying the hidden layers. The accuracy curves were generated for the six cases for the different parameter using CNN MNIST digit dataset.

The six cases perform differently because of the various combinations of hidden layers. The layers were taken randomly in a periodic sequence so that each case behaves differently during the experiment. The maximum and minimum accuracies were observed for different hidden layers variation with a batch size of 100. Among all the observation, the maximum accuracy in the performance was found 99.21% for 15 epochs in case 2 (Conv1, pool1, Conv2, pool2 with 2 dropouts). In digit recognition, this type of higher accuracy will cooperate to speed up the performance of the machine more adequately. However, the minimum accuracy among all observation in the performance was found 97.07% in case 6 (Conv1, pool1, Conv2, pool2 with 1 dropout). Moreover, among all the cases, the total highest test loss is approximately 0.049449 found in case 3 without dropout and the total lowest test loss is approximately 0.026303 found in case 2 with dropout. This low loss will provide CNN better performance to attain better image resolution and noise processing. In the future, we plan to observe the variation in the overall classification accuracy by varying the number of hidden layers and batch size.

## REFERENCES

[1] Y. LeCun *et al.*, "Backpropagation applied to handwritten zip code recognition," *Neural computation,* vol. 1, no. 4, pp. 541-551, 1989.

[2] A. Krizhevsky, I. Sutskever, and G. E. Hinton, "Imagenet classification with deep convolutional neural networks," in *Advances in neural information processing systems*, 2012, pp. 1097-1105.

[3] D. Hubel and T. Wiesel, "Aberrant visual projections in the Siamese cat," *The Journal of physiology,* vol. 218, no. 1, pp. 33-62, 1971.

[4] Y. LeCun, Y. Bengio, and G. Hinton, "Deep learning," *nature,* vol. 521, no. 7553, p. 436, 2015.

[5] D. Cireşan, U. Meier, and J. Schmidhuber, "Multi-column deep neural networks for image classification," *arXiv preprint arXiv:1202.2745,* 2012.

[6] K. Fukushima and S. Miyake, "Neocognitron: A self-organizing neural network model for a mechanism of visual pattern recognition," in *Competition and cooperation in neural nets*: Springer, 1982, pp. 267-285.

[7] Y. LeCun *et al.*, "Handwritten digit recognition with a back-propagation network," in *Advances in neural information processing systems*, 1990, pp. 396-404.

[8] Y. LeCun, L. Bottou, Y. Bengio, and P. Haffner, "Gradient-based learning applied to document recognition," *Proceedings of the IEEE,* vol. 86, no. 11, pp. 2278-2324, 1998.

[9] R. Hecht-Nielsen, "Theory of the backpropagation neural network," in *Neural networks for perception*: Elsevier, 1992, pp. 65-93.

[10] Y. LeCun, "LeNet-5, convolutional neural networks," *URL: http://yann. lecun. com/exdb/lenet,* vol. 20, 2015.

[11] S. J. Russell and P. Norvig, *Artificial intelligence: a modern approach*. Malaysia; Pearson Education Limited, 2016.

[12] S. Haykin, "Neural networks: A comprehensive foundation: MacMillan College," *New York,* 1994.

[13] K. B. Lee, S. Cheon, and C. O. Kim, "A convolutional neural network for fault classification and diagnosis in semiconductor manufacturing processes," *IEEE Transactions on Semiconductor Manufacturing,* vol. 30, no. 2, pp. 135-142, 2017.

[14] K. G. Pasi and S. R. Naik, "Effect of parameter variations on accuracy of Convolutional Neural Network," in *2016 International Conference on Computing, Analytics and Security Trends (CAST)*, 2016, pp. 398-403: IEEE.

[15] D. C. Ciresan, U. Meier, J. Masci, L. M. Gambardella, and J. Schmidhuber, "Flexible, high performance convolutional neural networks for image classification," in *Twenty-Second International Joint Conference on Artificial Intelligence*, 2011.

[16] K. Isogawa, T. Ida, T. Shiodera, and T. Takeguchi, "Deep shrinkage convolutional neural network for adaptive noise reduction," *IEEE Signal Processing Letters,* vol. 25, no. 2, pp. 224-228, 2018.

[17] C. C. Park, Y. Kim, and G. Kim, "Retrieval of sentence sequences for an image stream via coherence recurrent convolutional networks," *IEEE transactions on pattern analysis and machine intelligence,* vol. 40, no. 4, pp. 945-957, 2018.

[18] Y. Yin, J. Wu, and H. Zheng, "Ncfm: Accurate handwritten digits recognition using convolutional neural networks," in *2016 International Joint Conference on Neural Networks (IJCNN)*, 2016, pp. 525-531: IEEE.

[19] L. Xie, J. Wang, Z. Wei, M. Wang, and Q. Tian, "Disturblabel: Regularizing cnn on the loss layer," in *Proceedings of the IEEE Conference on Computer Vision and Pattern Recognition*, 2016, pp. 4753-4762.

[20] A. Tavanaei and A. S. Maida, "Multi-layer unsupervised learning in a spiking convolutional neural network," in *2017 International Joint Conference on Neural Networks (IJCNN)*, 2017, pp. 2023-2030: IEEE.

[21] J. Jin, K. Fu, and C. Zhang, "Traffic sign recognition with hinge loss trained convolutional neural networks," *IEEE Transactions on Intelligent Transportation Systems,* vol. 15, no. 5, pp. 1991-2000, 2014.

[22] M. Wu and Z. Zhang, "Handwritten digit classification using the mnist data set," *Course project CSE802: Pattern Classification & Analysis,* 2010.

[23] Y. Liu and Q. Liu, "Convolutional neural networks with large-margin softmax loss function for cognitive load recognition," in *2017 36th Chinese Control Conference (CCC)*, 2017, pp. 4045-4049: IEEE.

[24] Y. LeCun, "The MNIST database of handwritten digits," *http://yann. lecun. com/exdb/mnist/,* 1998.

[25] M. A. Nielsen, *Neural networks and deep learning*. Determination press USA, 2015.

[26] B. Zhang and S. N. Srihari, "Fast k-nearest neighbor classification using cluster-based trees," *IEEE Transactions on Pattern analysis and machine intelligence,* vol. 26, no. 4, pp. 525-528, 2004.

[27] E. Kussul and T. Baidyk, "Improved method of handwritten digit recognition tested on MNIST database," *Image and Vision Computing,* vol. 22, no. 12, pp. 971-981, 2004.

[28] R. B. Arif, M. A. B. Siddique, M. M. R. Khan, and M. R. Oishe, "Study and Observation of the Variations of Accuracies for Handwritten Digits Recognition with Various Hidden Layers and Epochs using Convolutional Neural Network," in *2018 4th International Conference on Electrical Engineering and Information & Communication Technology (iCEEiCT)*, 2018, pp. 112-117: IEEE.

[29] A. B. Siddique, M. M. R. Khan, R. B. Arif, and Z. Ashrafi, "Study and Observation of the Variations of Accuracies for Handwritten Digits Recognition with Various Hidden Layers and Epochs using Neural Network Algorithm," in *2018 4th International Conference on Electrical Engineering and Information & Communication Technology (iCEEiCT)*, 2018, pp. 118-123: IEEE.